\documentclass[lettersize,journal]{IEEEtran}
\usepackage{amsmath,amsfonts}
\usepackage{algorithmic}
\usepackage{algorithm}
\usepackage{array}
\usepackage[caption=false,font=normalsize,labelfont=sf,textfont=sf]{subfig}
\usepackage{textcomp}
\usepackage{stfloats}
\usepackage{url}
\usepackage{verbatim}
\usepackage{graphicx}
\usepackage{cite}
\usepackage{multirow}
\usepackage{makecell}
\usepackage{bm}
\usepackage{hyperref}
\usepackage{cleveref}
\usepackage{caption}
\begin{document}

\title{Biomechanics-Guided Residual Approach to Generalizable Human Motion Generation and Estimation}

\author{Zixi~Kang,
Xinghan~Wang,
and~Yadong~Mu% <-this % stops a space
% \thanks{Zixi Kang, Xinghan Wang and Yadong Mu are with the Wangxuan Institute of Computer Technology, Peking University, Beijing 100871, China (e-mail: forever.kzx0713@stu.pku.edu.cn, xinghan\_wang@pku.edu.cn, myd@pku.edu.cn).}

% \thanks{Yadong Mu is the corresponding author.}
}

% \maketitle


\twocolumn[{
\renewcommand\twocolumn[1][]{#1}
\maketitle

\vspace{-25pt}

\begin{center}
    \centering
    \begin{minipage}{0.61\linewidth}
        \centering
        \includegraphics[width=\linewidth]{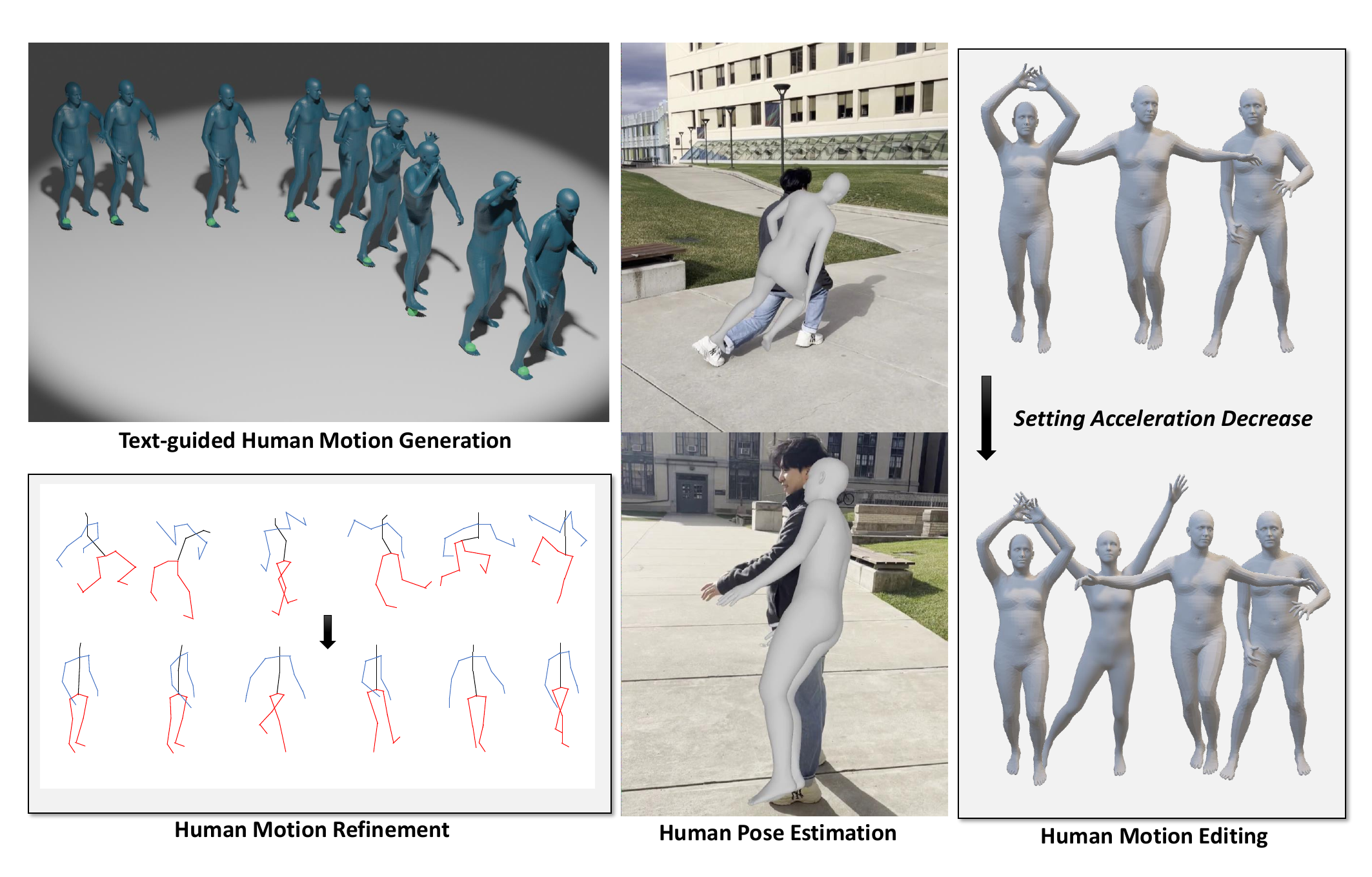}
        \vspace{2pt}
        \small (a) Results across multiple tasks
    \end{minipage}
    \hfill
    % 子图 b
    \begin{minipage}{0.38\linewidth}
        \centering
        \includegraphics[width=\linewidth]{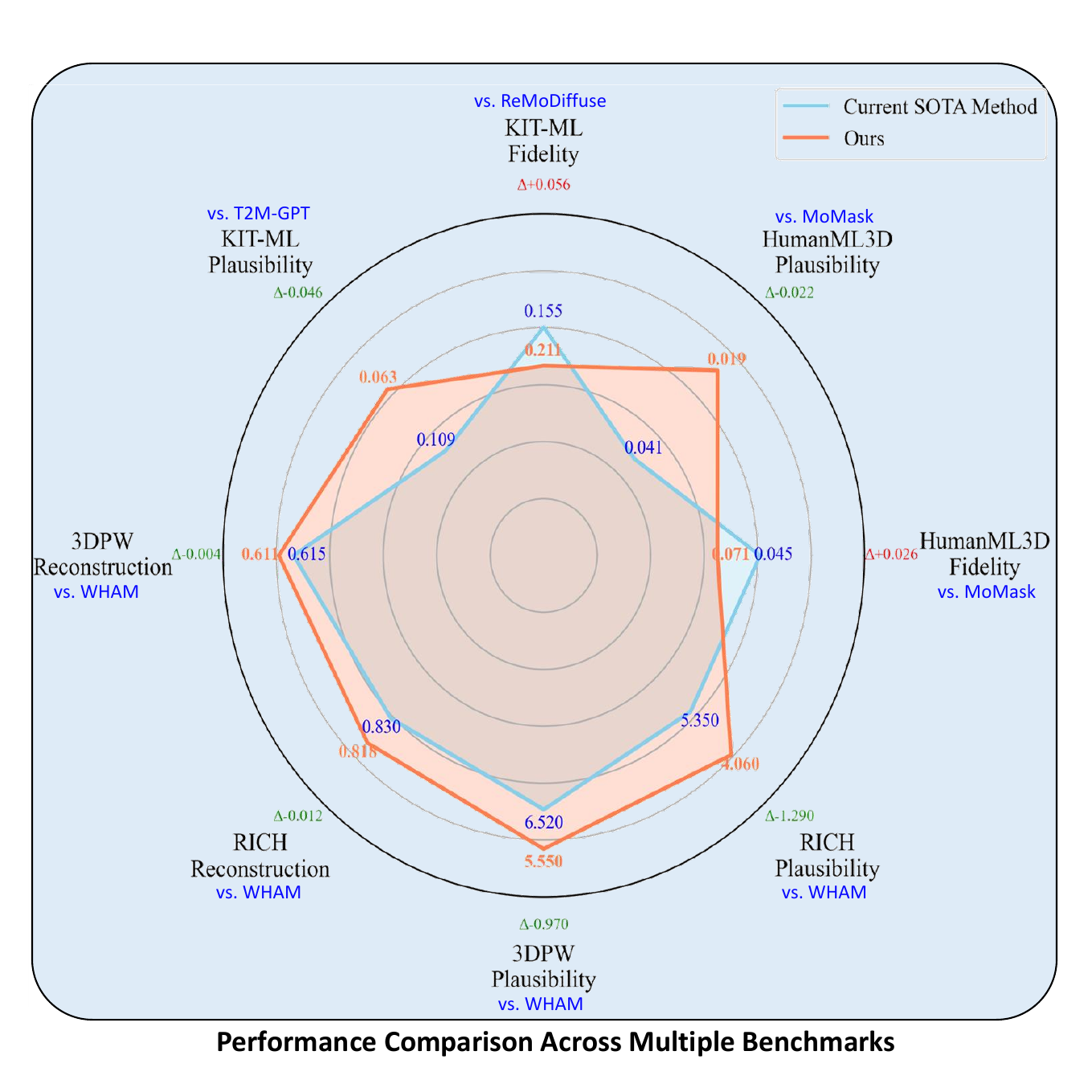}
        \vspace{2pt}
        \small (b) Quantitative comparison
    \end{minipage}

    \vspace{-5pt}

    \captionof{figure}{\small
    (a) Demonstrates our method’s results across multiple tasks, including motion generation, motion refinement, pose estimation, and motion editing. (b) Shows a quantitative comparison of different metrics against current state-of-the-art methods. Our proposed \textbf{BioVAE} framework demonstrates strong generalization across these tasks. Compared to state-of-the-art methods, our approach generates motions with significantly reduced foot floating, penetration, and clipping artifacts, as well as improved temporal smoothness. \textbf{Notably, BioVAE achieves a clear improvement in physical plausibility metrics, setting a new benchmark for biomechanically realistic motion synthesis.}
    }
    \label{fig:mainpage}
\end{center}}]

\begingroup
    \renewcommand\thefootnote{}\footnotetext{
    Zixi Kang, Xinghan Wang and Yadong Mu are with the Wangxuan Institute of Computer Technology, Peking University, Beijing 100871, China (e-mail: forever.kzx0713@stu.pku.edu.cn, xinghan\_wang@pku.edu.cn, myd@pku.edu.cn).
    
    Yadong Mu is the corresponding author.
    }
\endgroup

\begin{abstract}
Human pose, action, and motion generation are critical for applications in digital humans, character animation, and humanoid robotics. However, many existing methods struggle to produce physically plausible movements that are consistent with biomechanical principles. Although recent autoregressive and diffusion models deliver impressive visual quality, they often neglect key biodynamic features and fail to ensure physically realistic motions. Reinforcement Learning (RL) approaches can address these shortcomings but are highly dependent on simulation environments, limiting their generalizability. To overcome these challenges, we propose BioVAE, a biomechanics-aware framework with three core innovations: (1) integration of muscle electromyography (EMG) signals and kinematic features with acceleration constraints to enable physically plausible motion without simulations; (2) seamless coupling with diffusion models for stable end-to-end training; and (3) biomechanical priors that promote strong generalization across diverse motion generation and estimation tasks. Extensive experiments demonstrate that BioVAE achieves state-of-the-art performance on multiple benchmarks, bridging the gap between data-driven motion synthesis and biomechanical authenticity while setting new standards for physically accurate motion generation and pose estimation.
\end{abstract}


\section{Introduction}
\label{sec:intro}

Human pose, action, and motion generation have wide-ranging applications in motion capture, animation, and robotics. In recent years, there has been a growing focus on generating human motion conditioned on image, language, sound, and motion trajectories~\cite{tm2d,Petrovich2024CVPR,opendance,wham,semantic,GestureDiffuCLIP,EHPE,CMQ,EmotionGesture,MANet,Text2Avatar,StridedTransformer}.

However, studies in biomechanics and animation\cite{fpsmetric, motioneval} have shown that current methods fail to capture the distinct characteristics of human motion, resulting in generated sequences that often deviate significantly from natural movement, with unnatural rhythms and unrealistic joint configurations. Moreover, the lack of explicit modeling of biodynamic features prevents these models from capturing critical information such as acceleration patterns and muscle activation levels, often resulting in motions that appear stiff or violate physical laws. Furthermore, existing evaluation metrics also overlook the quantitative assessment of physical plausibility and motion velocity, while current approaches struggle to control movement amplitude and speed precisely. These limitations severely undermine the practical value of motion generation in professional applications that demand high biomechanical precision.

To address these challenges, we propose \textbf{BioVAE}, a biomechanics-aware framework that seamlessly integrates musculoskeletal dynamics into diffusion processes. Our key innovation lies in embedding Euler-Lagrange equations and EMG signals into the VAE architecture, while connecting the VAE residuals to the diffusion process to enable end-to-end, generalizable motion generation. Furthermore, since current evaluation protocols largely overlook biomechanical aspects, we establish a comprehensive set of metrics which emphasizes physical realism, combining biomechanical validity measures (\emph{e.g.}, motion smoothness, foot sliding rate) with traditional generation metrics such as FID and diversity. Extensive experiments on HumanML3D\cite{humanml3d}, KIT-ML\cite{kitml}, 3DPW\cite{3DPW}, and RICH\cite{RICH} demonstrate that BioVAE achieves significant improvements over existing methods.

The main contributions of this paper are as follows.
\begin{itemize}
\item We propose \textbf{BioVAE}, a residual approach based on Euler-Lagrange equations that can be jointly trained with diffusion models in a self-supervised setting.
\item We introduce EMG signals and acceleration-based physical supervision, leveraging additional biomechanical priors to enhance the model’s capability.
\item We design extensive experiments across different tasks to evaluate both reconstruction quality and physical plausibility, demonstrating the strong generalizability of our framework.
\end{itemize}

\section{Related Work}
\label{sec:related}

\subsection{Physically plausible motion processing.} 
In motion generation, Reinforcement Learning (RL) has been utilized to fine-tune models for physical plausibility~\cite{aligning,phc,pbc}. For example, ReinDiffuse~\cite{reindiffuse} and PhysDiff~\cite{physdiff} extend MDM~\cite{MDM} to improve motion quality. However, these RL-based methods suffer from two major limitations: (1) they lack the ability of generalization and require task-specific, dataset-specific, and model-specific reward designs; and (2) they depend heavily on simulation environments, making the training process a black box and preventing end-to-end optimization. 

Alternatively, some methods incorporate auxiliary constraints during training~\cite{omnicontrol,gmd,tracepace}; for instance, OmniControl~\cite{omnicontrol} leverages spatial signals to guide generation. However, since these approaches mainly focus on enabling multimodal and editable motion generation instead of improving motion quality, they remain prone to challenges like limited precision and excessive jitter.  Although techniques like Gaussian filtering can suppress high-frequency noise, they often degrade motion precision. 

Consequently, several studies~\cite{gaussianfilter,tcmr,skeletor,smoothnet,temporalrefinement,attentionrefinement,flexmotion,rfc} have explored pose refinement techniques to reduce temporal instability. For example, SmoothNet~\cite{smoothnet} introduces a parametric approach to optimize temporal coherence and substantially improve motion quality. Nevertheless, these methods do not address generalization and cannot be seamlessly integrated with diffusion models or large language models (LLMs), failing to leverage the powerful generative capabilities of such backbones and exhibiting inherent limitations.

\subsection{Diffusion models.} 
Diffusion models have achieved remarkable success in image generation, with numerous approaches leveraging language guidance to produce high-quality results~\cite{stablediffusion,imagen,controlnet}. Among them, Denoising Diffusion Probablistic Models (DDPM)~\cite{ddpm} has become the most widely adopted sampling method, delivering impressive results. Based on DDPM, DDIM~\cite{ddim} reduces inference steps and allows deterministic sampling by adjusting the variance schedule. The introduction of Classifier Guidance (CG)\cite{cg} extended diffusion models to conditional generation tasks, while Classifier-Free Guidance (CFG)\cite{cfg} further streamlined this process by replacing CG and improving the efficiency of the guidance during training. Inspired by these advancements, we observe that incorporating domain-specific knowledge is conceptually aligned with the mechanisms of CG and CFG, thereby motivating a re-derivation of the diffusion process.

In parallel, recent studies have addressed the diffusion inversion problem, which map real images to latent space and supports various image editing applications~\cite{nulltextinversion,edict,stylediffusion,directinversion,inst,diffedit}. However, diffusion inversion techniques have rarely been explored beyond image generation. This work is the first to apply DDIM inversion to motion generation tasks.

\subsection{Text-driven human motion generation.}
Early approaches to text-driven human motion generation conceptualized motion as a form of language, adopting autoregressive methods commonly used in language models~\cite{language2pose,text2action,temos,teach}. Recognizing the potential of large language models, subsequent studies improved motion embeddings to enable joint learning of motion and text, yielding promising results~\cite{t2m,tm2t,t2mgpt,motionlanguage,momask,mmm}.

MDM~\cite{MDM} pioneered the application of diffusion models to human motion generation, while MLD~\cite{MLD}, inspired by VAE~\cite{vae}, introduced a diffusion process within the latent space. MotionDiffuse~\cite{motiondiffuse} established a larger model framework, using cross-attention to link motion and text. Building on MotionDiffuse, ReMoDiffuse~\cite{remodiffuse} incorporated retrieval-based motion databases and supplementary knowledge, resulting in significant performance improvements. MoFusion~\cite{mofusion} and StableMoFusion~\cite{stablemofusion}, drawing inspiration from Stable Diffusion~\cite{stablediffusion}, utilized the UNet~\cite{unet} architecture without relying on external knowledge, achieving state-of-the-art results. To enable our \textbf{BioVAE} to deeply integrate with the diffusion process, we formulate BioVAE as a self-supervised residual connection, re-derive the probability formulation of the diffusion process, and redesign the loss function to achieve stable end-to-end training of the two components.

\section{Method}
\label{sec:method}

\begin{figure*}[]
    \centering
    \includegraphics[width=0.88\linewidth]{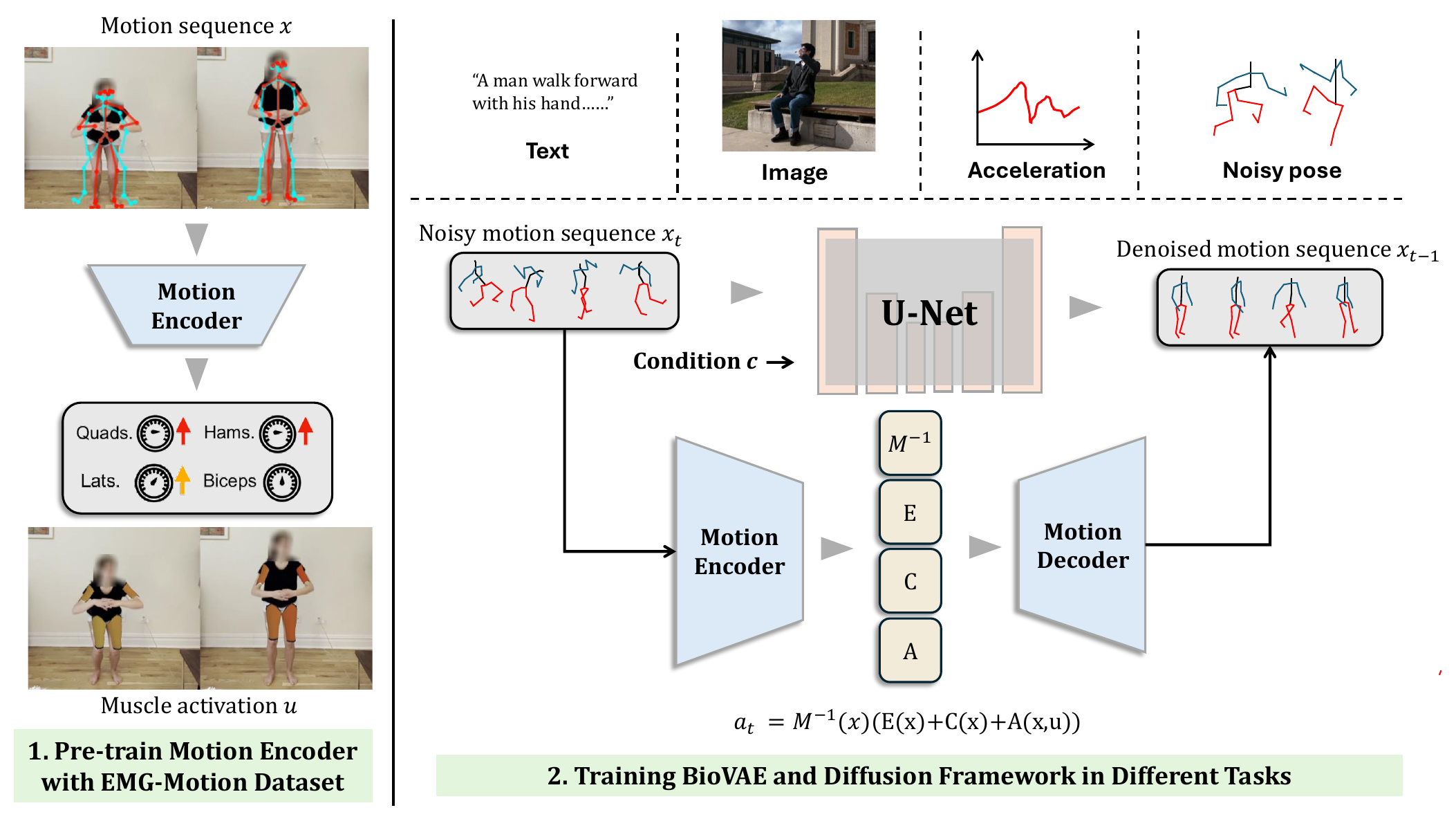}
    \caption{
    \textbf{\small Overview of BioVAE.} We employ the MIA dataset~\cite{musclesinaction} and adopt a training strategy that learns the mapping from motion sequences to EMG signals within the motion encoder. In the latent space, a neural network predicts four intermediate variables: \(M^{-1}\), \(E\), \(C\), and \(A\), which are then combined through the Euler-Lagrange equations to reconstruct the joint local rotational accelerations. The motion decoder iteratively generates residuals from the acceleration information to guide the UNet’s generation process. \textbf{BioVAE} can be applied to a variety of tasks and modalities, such as text-to-motion, pose estimation, motion editing, and motion refinement, demonstrating strong generalization across tasks while significantly enhancing the physical plausibility of the generated motions.  
    }
    \label{fig:method}
    \vspace{-10pt}
\end{figure*}

Existing methods fail to capture the unique characteristics of human motion due to the absence of biomechanical constraints and the lack of rigid body modeling. In this section, we present our \textbf{BioVAE} framework, which integrates biomechanical principles and rigid body dynamics into the network architecture, providing a diffusion-based model that achieves state-of-the-art performance in multiple tasks. We begin in \Cref{subsec:pre} by introducing the preliminaries. \Cref{subsec:net} details the architecture of the BioVAE network. In \Cref{subsec:train}, we derive the joint distribution formulation that allows end-to-end training of BioVAE with diffusion through residual connections. Finally, \Cref{subsec:inverse} describes our diffusion inversion strategy, showing how BioVAE supports motion refinement, motion editing, and pose estimation.

\subsection{Preliminaries}
\label{subsec:pre}

Human motion generation produces a motion sequence \( \Theta \in \mathbb{R}^{T \times D} \), where \( T \) denotes the length of the sequence and \( D \) is the dimension of the joint representation. This motion sequence can be transformed into a joint coordinate representation \( X \in \mathbb{R}^{T \times 3d} \), where \( d \) represents the number of joint nodes. By modeling the human body as an articulated rigid-body system, the Euler-Lagrange equation can be formulated as:
\begin{equation}
F = M \cdot \ddot{X_t} + C,
\end{equation}
where \( F \) represents the internal and external contact forces, \( M \) is the inertia matrix, and \( C \) captures the combined effect of the Coriolis force and gravity. The term \( F \) can be decomposed into internal force \( I \) and the external contact force \( E \). According to biomechanical principles, the internal force \( I \) is generated by muscle movements and can be expressed as:
\begin{equation}
I = R \cdot A(u),
\end{equation}
where \( R \) is the muscle rotation matrix, \( A(u) \) is the muscle activation force, and \( u \) is the muscle activation coefficient. Using this decomposition, the Euler-Lagrange equation for human motion can be rewritten as \Cref{eq:lagrange}.

\begin{equation}
\ddot{X_t} = M^{-1}( E + C + R \cdot A(u)).
    \label{eq:lagrange}
\end{equation}

\subsection{BioVAE Architecture}
\label{subsec:net}

Compared to conventional neural networks, a network based on the Euler-Lagrange equation provides a stronger prior, as it can inherently capture acceleration dynamics while leveraging constraints to reduce the complexity of the problem. In this section, we introduce how to design a VAE architecture guided by the Euler-Lagrange equation.  

Following the representation in the MIA dataset, there exists a learnable mapping between human poses and the activation coefficients of the primary muscles \( u \in \mathbb{R}^8 \), defined as \( u = u(X_{:t}) \). The muscle rotation matrix \( R \) depends solely on the rotational component of the pose, while the inertia matrix \( M \) is an intrinsic property of the body determined only by position. The Coriolis force and gravity can be expressed as follows.  
\begin{equation}
C = m\omega^2r + 2m\omega \times \dot{X}_t - \nabla U_{gravity}(X_t),
\label{eq:coriolis}
\end{equation}  
which are likewise pose-dependent. The external contact force \( F \), including the elastic and resistive forces, depends on both position and velocity. Using this formulation, Equation~\eqref{eq:lagrange} can be rewritten as a function of \( X_{:t} \) in Equation~\eqref{eq:vae-lagrange}:  
\begin{equation}
\ddot{X}_t = M^{-1}(X_{:t}) \left( F(X_{:t}) + C(X_{:t}) + R(X_{:t}) \cdot A(u) \right).
\label{eq:vae-lagrange}
\end{equation}  

We implement three neural networks to approximate \( M^{-1} \), \( F \), and \( C \), where \( M^{-1} \) is restricted to be symmetric. The term \( R \cdot A(u) \) is modeled using cross-attention, allowing four lightweight networks to predict the acceleration of the pose effectively. Once acceleration \( \ddot{X}_t \) is predicted, we employ an autoregressive decoder to reconstruct the residual between predicted \( X_t \) and the ground truth. This enables optimization at each training time step while ensuring temporal consistency and biomechanical fidelity.

\begin{table*}[htbp]
\centering
\caption{\small \textbf{Quantitative evaluation on the HumanML3D and KIT-ML test set.} The confidence interval is indicated by$\pm$. \textbf{Bold} face indicates the best result, while underscore refers to the second best. We reimplemented StableMoFusion based on the official code.}
\label{tab:evaluator}
\setlength{\tabcolsep}{0.75mm}
\renewcommand\arraystretch{1.2}
{
\begin{tabular}{llcccccccc}
\hline
\multirow{2}{*}{Dataset} & \multirow{2}{*}{Methods} & \multicolumn{3}{c}{R Precision$\uparrow$} & \multirow{2}{*}{FID$\downarrow$} & \multirow{2}{*}{MM Dist$\downarrow$} & \multirow{2}{*}{Diversity$\rightarrow$} & \multirow{2}{*}{MultiModality$\uparrow$} \\
& &  Top 1 & Top 2 & Top 3 \\
\hline
\multirow{8}{*}{\makecell[c]{KIT-\\ML}} 
% & GT             & $0.424^{\pm.005}$ & $0.649^{\pm.006}$ & $0.779^{\pm.006}$ & $0.031^{\pm.004}$ & $2.788^{\pm.012}$ & $11.08^{\pm.097}$ & - \\
& MDM            & $0.164^{\pm.004}$ & $0.291^{\pm.004}$ & $0.396^{\pm.004}$ & $0.497^{\pm.021}$ & $9.191^{\pm.022}$ & $10.85^{\pm.109}$ & \underline{$1.91^{\pm.214}$} \\
& MLD            & $0.390^{\pm.008}$ & $0.609^{\pm.008}$ & $0.734^{\pm.027}$ & $0.404^{\pm.027}$ & $3.204^{\pm.027}$ & $10.80^{\pm.117}$ & \bm{$2.19^{\pm.071}$} \\
& T2M            & $0.370^{\pm.005}$ & $0.569^{\pm.007}$ & $0.693^{\pm.007}$ & $2.770^{\pm.109}$ & $3.401^{\pm.008}$ & $10.91^{\pm.119}$ & $1.48^{\pm.065}$ \\
& MotionDiffuse  & $0.417^{\pm.004}$ & $0.621^{\pm.004}$ & $0.739^{\pm.004}$ & $1.954^{\pm.062}$ & $2.958^{\pm.005}$ & \bm{$11.10^{\pm.143}$} & $0.73^{\pm.013}$ \\
& T2M-GPT        & $0.416^{\pm.006}$ & $0.627^{\pm.006}$ & $0.745^{\pm.006}$ & $0.514^{\pm.029}$ & $3.007^{\pm.023}$ & $10.92^{\pm.108}$ & $1.57^{\pm.039}$ \\
& ReMoDiffuse    & $0.427^{\pm.014}$ & $0.641^{\pm.004}$ & $0.765^{\pm.055}$ & \bm{$0.155^{\pm.006}$} & $2.814^{\pm.012}$ & $10.80^{\pm.105}$ & $1.24^{\pm.028}$ \\
& StableMoFusion & \underline{$0.445^{\pm.006}$} & \underline{$0.660^{\pm.005}$} & \underline{$0.782^{\pm.004}$} & $0.258^{\pm.029}$ & $2.832^{\pm.014}$ & $10.94^{\pm.077}$ & $1.36^{\pm.062}$ \\
& MoMask & $0.433^{\pm .007}$ & $0.656^{\pm .005}$ & $0.781^{\pm .005}$ & \underline{$0.204^{\pm .011}$} & \underline{$2.779^{\pm .022}$} & - & $1.13^{\pm .043}$ \\
\cline{2-9}
& Ours           & \bm{$0.448^{\pm.008}$} & \bm{$0.666^{\pm.005}$} & \bm{$0.788^{\pm.005}$} & $0.211^{\pm.101}$ & \bm{$2.772^{\pm.017}$} & \underline{$11.11^{\pm.094}$} & $1.38^{\pm.050}$ \\
\hline
\multirow{8}{*}{\makecell[c]{Human\\ML3D}} 
% & GT             & $0.511^{\pm.003}$ & $0.703^{\pm.003}$ & $0.797^{\pm.002}$ & $0.002^{\pm.000}$ & $2.974^{\pm.008}$ & $9.503^{\pm.065}$ & - \\
& MDM            & $0.320^{\pm.005}$ & $0.498^{\pm.004}$ & $0.611^{\pm.007}$ & $0.544^{\pm.044}$ & $5.566^{\pm.027}$ & \bm{$9.559^{\pm.086}$} & \bm{$2.799^{\pm.072}$} \\
& MLD            & $0.481^{\pm.003}$ & $0.673^{\pm.003}$ & $0.772^{\pm.002}$ & $0.473^{\pm.013}$ & $3.196^{\pm.010}$ & $9.724^{\pm.082}$ & \underline{$2.413^{\pm.079}$} \\
& T2M            & $0.457^{\pm.002}$ & $0.639^{\pm.003}$ & $0.743^{\pm.003}$ & $1.067^{\pm.002}$ & $3.340^{\pm.008}$ & $9.188^{\pm.002}$ & $2.090^{\pm.083}$ \\
& MotionDiffuse  & $0.491^{\pm.001}$ & $0.681^{\pm.001}$ & $0.782^{\pm.001}$ & $0.630^{\pm.001}$ & $3.113^{\pm.001}$ & $9.410^{\pm.049}$ & $1.553^{\pm.042}$ \\
& T2M-GPT        & $0.491^{\pm.003}$ & $0.680^{\pm.003}$ & $0.775^{\pm.002}$ & $0.116^{\pm.004}$ & $3.118^{\pm.011}$ & \bm{$9.761^{\pm.081}$} & $1.856^{\pm.011}$ \\
& ReMoDiffuse    & $0.510^{\pm.005}$ & $0.698^{\pm.006}$ & $0.795^{\pm.004}$ & $0.103^{\pm.004}$ & $2.974^{\pm.016}$ & $9.018^{\pm.075}$ & $1.795^{\pm.043}$ \\
& StableMoFusion & \underline{$0.545^{\pm.003}$} & \underline{$0.743^{\pm.003}$} & \underline{$0.833^{\pm.002}$} & $0.133^{\pm.005}$ & \underline{$2.814^{\pm.008}$} & $9.704^{\pm.065}$ & $1.844^{\pm.060}$ \\
& MoMask & $0.521^{\pm .002}$ & $0.713^{\pm .002}$ & $0.807^{\pm .002}$ & $\bm{0.045}^{\pm .002}$ & $2.958^{\pm .008}$ & $9.644^{\pm .086}$ & $1.241^{\pm .040}$ \\
\cline{2-9}
& Ours           & \bm{$0.547^{\pm.003}$} & \bm{$0.743^{\pm.002}$} & \bm{$0.835^{\pm.002}$} & \underline{$0.071^{\pm.003}$} & \bm{$2.784^{\pm.008}$} & \underline{$9.567^{\pm.086}$} & $1.919^{\pm.063}$ \\
\hline
\end{tabular}}
\vspace{-10pt}
\end{table*}

\subsection{End-to-End Residual Training}
\label{subsec:train}

In previous approaches, optimization methods such as reinforcement learning (RL) or refinement~\cite{robotMDM,smoothnet} are often applied as post-processing on top of the backbone model. In contrast, we integrate BioVAE with diffusion to enable end-to-end training. Furthermore, we apply optimization at every diffusion training timestep, allowing our model to perform iterative refinement during inference, which substantially improves the quality of generation.  

Traditional RL and simulation-based methods~\cite{pbc,phc,physdiff} typically require a complete human motion sequence and thus cannot be integrated into the noisy training process of diffusion models. To address this, our network explicitly treats acceleration \(\ddot{X_t}\) as an intermediate variable. We introduce supervised constraints on the acceleration during the diffusion process to ensure that the predicted accelerations are physically meaningful, resulting in a more stable training process and faster convergence.  

Specifically, at each reverse diffusion step, we consider extending the distribution at step $t$, denoted as $\Theta^t$, to a joint distribution $S^t = \{\Theta^t, \ddot{X}^t\}$. According to DDPM theory, when performing diffusion on $S^t$, the model's loss function can be written as:
\begin{equation}
\begin{aligned}
&\min E_{Q(S^0 : S^T)}\left[ -\log P_\theta(S^0 | S^1) \right. \\
&+ \sum_{t=2}^T \text{KL}(Q(S^{t-1} | S^t, S^0) \| \left. P_\theta(S^{t-1} | S^t))\right].
\end{aligned}
\label{eq_diff}
\end{equation}

We argue that, during the denoising process, the quantities \( \Theta^{t-1}, \ddot{X}^{t-1} \) at step \( t-1 \) are unknown and do not satisfy the interpolation relationship. However, the known quantities \( \Theta^t, \ddot{X}^t \) at step \( t \) do satisfy this relationship. This leads to the following:
\begin{equation}
\begin{aligned}
&\log P_\theta(S^{t-1} | S^t) = \log P_\theta(\Theta^{t-1}, \ddot{X}^{t-1} | \Theta^t, \ddot{X}^t) \\
&= \log P_\theta(\Theta^{t-1}, \ddot{X}^{t-1} | \Theta^t) \\
&= \log P_\theta(\Theta^{t-1} | \ddot{X}^{t-1}, \Theta^t) + \log P_\theta(\ddot{X}^{t-1} | \Theta^t).
\end{aligned}
\label{eq:physnet}
\end{equation}

In the expression in \Cref{eq:physnet}, \( \log P_\theta(\Theta^{t-1} | \ddot{X}^{t-1}, \Theta^t) \) resembles the condition $c$ in CG and CFG, while \( \log P_\theta(\ddot{X}^{t-1} | \Theta^t) \) can be obtained using the VAE architecture we propose. Thus, during training, we set \( \ddot{X}^{t-1} = \text{VAE}(\Theta^t) \), and the predicted acceleration is compared to the ground truth acceleration, introducing a physics loss \( L_{\text{phys}} = \text{MSE}(\text{VAE}(\Theta^t), \ddot{X_{\text{gt}}}^t) \). To further enhance acceleration accuracy, we design the following additional loss term:
\begin{equation}
L_{\text{accel}} = \text{MSE}(\ddot{X}^t, \ddot{X_{\text{gt}}}^t).
\end{equation}

Thus, the total loss function during training is written as:
\begin{equation}
L = L_{\text{diffusion}} + L_{\text{accel}} + L_{\text{phys}}.
\end{equation}

\subsection{Generation with Diffusion Inversion}
\label{subsec:inverse}

\begin{figure}[htbp]
    \centering
    \includegraphics[width=\linewidth]{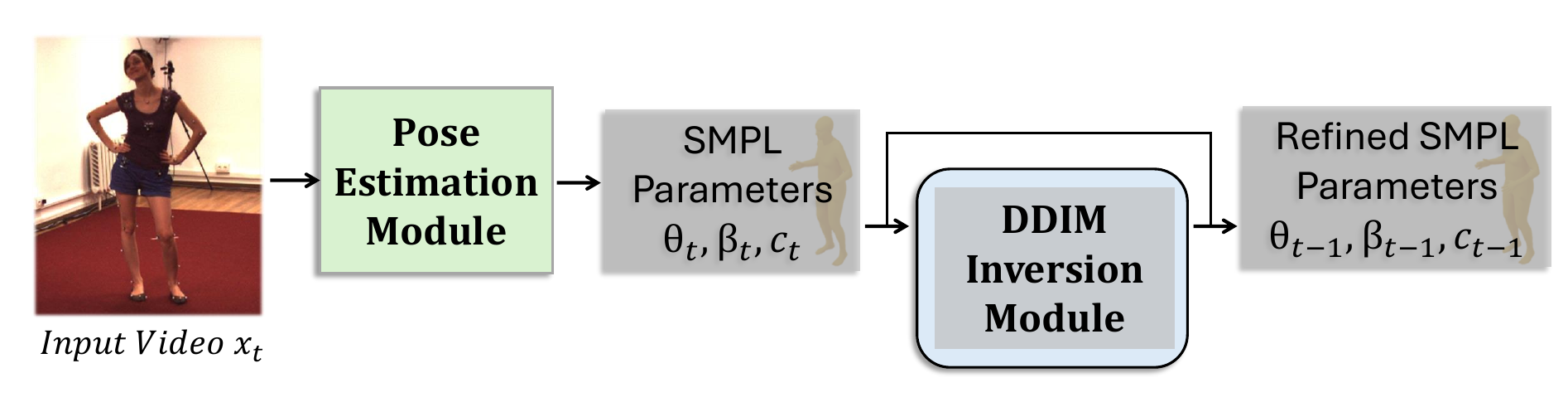}
    \caption{\small
    \textbf{Estimation, refinement, and editing using DDIM Inversion.} We incorporate the BioVAE-Diffusion framework as a residual module and apply diffusion inversion to enable pose estimation, motion refinement, and motion editing tasks. By adopting an \textit{inversion-then-forward} pipeline, our approach significantly improves the smoothness of generated SMPL sequences while enhancing the overall generation quality.
    }
    \label{fig:inversion}
    \vspace{-8pt}
\end{figure}

BioVAE can be readily adapted to pose estimation, motion refinement, and motion editing tasks. The key insight is that our BioVAE-Diffusion framework is capable of generating physically plausible motions starting from latent noise. Given the initial estimates of human motion, we first employ diffusion inversion~\cite{edict,directinversion}—a technique that reverses the usual diffusion process—to recover the latent noise corresponding to the motion sequence. We then feed this noise into our model, which injects biomechanical constraints to produce motion sequences that are more authentic compared to the original inputs.  

Specifically, given the initial SMPL~\cite{SMPL} parameters \(\mathbf{x}_0\), we first invert them to timestep \(t\) using DDIM sampling:  
\begin{equation}
\begin{aligned}
\mathbf{x}_t = \sqrt{\alpha_t}\mathbf{x}_0 + \sqrt{1 - \alpha_t}\epsilon_{\theta}(\mathbf{x}_0, t),
\end{aligned}
\label{eq:inversion}
\end{equation}
where \(t\) controls the balance between information preservation and noise introduction. The inverted latent \(\mathbf{x}_t\) then undergoes forward BioVAE-Diffusion to produce the refined motion:  
\begin{equation}
\begin{aligned}
\hat{\mathbf{x}_0} = f_{\theta}(\mathbf{x}_t, t).
\end{aligned}
\label{eq:forward_inversion}
\end{equation}  

Crucially, our parametric implementation inherits the physics constraints from \Cref{subsec:net,subsec:train} without requiring additional tuning, demonstrating the strong generalization capability of our method across multiple tasks.

\begin{table*}[thbp]
\centering
\caption{\small \textbf{Quantitative evaluation of physical plausible metrics on the HumanML3D and KIT-ML test set.} All experiments were repeated 20 times with different random seeds, and the mean results are reported in the tables. The results for T2M, and MotionDiffuse on HumanML3D is sourced from ReinDiffuse~\cite{reindiffuse}.}
\label{tab:physics_evaluator}
\setlength{\tabcolsep}{1.4mm}
\renewcommand\arraystretch{1.1}
{
\begin{tabular}{llccccccccc}
\hline
Dataset & Method & \makecell[c]{VC$\rightarrow$} & \makecell[c]{AC$\rightarrow$} & \makecell[c]{Speed$\rightarrow$} & \makecell[c]{Smoothness$\rightarrow$} & \makecell[c]{Float$\rightarrow$} & \makecell[c]{Foot Skating$\rightarrow$} & \makecell[c]{Penetrate$\rightarrow$} & \makecell[c]{Clip$\rightarrow$} \\
\hline
\multirow{7}{*}{\makecell[c]{KIT-\\ML}} 
& Real Motions              & 0.352 & 1.379 & 0.926 & 3.06  & 0.550 & 0.309 & 0.208      & 0.001 \\
\cline{2-10}
% & MDM             & -     & -     & -     & -     & -     & -     & -          & -     \\
& T2M             & $0.270$ & $1.519$ & $0.748$ & $4.90$  & $0.442$ & $0.324$ & $0.180$ & $\bm{0.001}$ \\
& MotionDiffuse   & $0.258$ & $1.861$ & $1.177$ & $1.92$ & $0.136$ & $0.808$ & $0.490$ & $0.005$ \\
& T2M-GPT         & $0.238$ & $\bm{1.420}$ & $0.832$ & $3.97$  & $0.503$ & $0.267$ & $0.188$ & $0.002$ \\
& ReMoDiffuse     & $2.826$ & $7.993$ & $1.415$ & $0.29$ & $0.254$ & $0.789$ & $\bm{0.199}$ & $0.006$ \\
& StableMoFusion  & $0.684$ & $2.496$ & $0.986$ & $1.80$ & $0.496$ & $0.315$ & $0.255$ & $0.002$ \\
\cline{2-10}
& Ours            & $\bm{0.426}$ & $1.736$ & $\bm{0.965}$ & $\bm{2.25}$ & $\bm{0.522}$ & $\bm{0.306}$ & $0.237$ & $0.003$ \\
\hline
\multirow{7}{*}{\makecell[c]{Human\\ML3D}} 
& Real Motions              & 0.124 & 0.682 & 0.358 & 2.60  & 0.205 & 0.057 & 0.000      & 0.000 \\
\cline{2-10}
% & MDM             & -     & -     & -     & -     & -     & 0.102 & 0.048      & 0.014 \\
& T2M             & $0.326$ & $2.826$ & $0.301$ & $1.13$ & $0.110$ & $0.217$ & $0.235$      & $0.263$ \\
& MotionDiffuse   & $0.082$ & $0.963$ & $0.332$ & $1.17$ & $0.031$ & $0.426$ & $0.386$      & $0.217$ \\
& T2M-GPT         & $0.178$ & $1.459$ & $\bm{0.354}$ & $1.28$ & $0.192$ & $0.052$ & $0.022$      & $\bm{0.001}$ \\
& ReMoDiffuse             & $2.353$ & $7.163$ & $0.605$ & $0.15$  & $0.040$ & $0.376$ & $0.019$      & $0.008$ \\
& StableMoFusion             & $0.292$ & $2.220$ & $0.448$ & $0.74$  & $0.234$ & $0.074$ & $0.043$     & $0.002$ \\
& MoMask             & $0.106$ & $1.048$ & $0.331$ & $\bm{1.89}$  & $0.261$ & $0.050$ & $0.013$     & $0.001$ \\
\cline{2-10}
& Ours            & $\bm{0.135}$ & $\bm{0.959}$ & $0.385$ & $1.45$ & $\bm{0.204}$ & $\bm{0.056}$ & $\bm{0.013}$ & $0.002$ \\
\hline
\end{tabular}}
\vspace{-5pt}
\end{table*}

\section{Experiments}
\label{sec:experiments}

\subsection{Datasets and Metrics}

We trained and evaluated our model on both the KIT-ML~\cite{kitml} and HumanML3D~\cite{humanml3d} datasets. The KIT-ML dataset contains 3,911 motions and 6,363 natural language annotations, while HumanML3D is derived from and reannotates the HumanAct12~\cite{action2motion} and AMASS~\cite{AMASS} datasets, comprising a total of 14,616 motion sequences and 44,970 descriptions.

We follow the evaluation metrics used in previous motion generation works~\cite{MDM,motiondiffuse,stablemofusion}, including Frechet Inception Distance (FID), Precision, Diversity, Multimodality, and Multi-Modal Distance. Additionally, we consider physics-related metrics such as skating ratio, ground floating, ground penetration, and foot clipping, which have also been used in prior studies~\cite{reindiffuse,omnicontrol,physdiff}. Furthermore, we introduce two novel metrics derived from Fourier analysis: Velocity Consistency (VC) and Acceleration Consistency (AC). Our motivation stems from the observation that while the Smoothness metric captures global temporal coherence, it overlooks the inherent kinematic relationship between stable body parts and agile extremities in human motion. For a motion sequence represented as $\Phi \in \mathbb{R}^{T \times 3d}$, the metrics are formally defined as:
\begin{equation}
\begin{aligned}
\text{VC} &= \frac{1 - \|\text{FFT}(\dot{\Phi})_{[:thre_{\text{low}}]}\| - \|\text{FFT}(\dot{\Phi})_{[thre_{\text{high}}:]}\|}{\|\text{FFT}(\dot{\Phi})_{[:thre_{\text{low}}]}\| + \|\text{FFT}(\dot{\Phi})_{[thre_{\text{high}}:]}\|}, \\
\text{AC} &= \frac{1 - \|\text{FFT}(\ddot{\Phi})_{[:thre_{\text{low}}]}\| - \|\text{FFT}(\ddot{\Phi})_{[thre_{\text{high}}:]}\|}{\|\text{FFT}(\ddot{\Phi})_{[:thre_{\text{low}}]}\| + \|\text{FFT}(\ddot{\Phi})_{[thre_{\text{high}}:]}\|}.
\end{aligned}
\label{eq:appendix1}
\end{equation}
Spectral ratio analysis enables VC and AC to effectively characterize overall motion patterns by precisely quantifying the balance between high-frequency components (associated with rapid limb movements) and low-frequency components (correlated with stable torso dynamics). This frequency-domain approach provides a more accurate assessment of motion physicality compared to traditional smoothness metrics.
Since the dataset includes motions like ``climbing stairs" (a foot is floating from the ground) and ``jumping while spinning" (abrupt acceleration), all these physics-based metrics should be made as consistent as possible with ground-truth instead of being blindly minimized to meet the characteristics of dataset.

\subsection{Evaluation Metrics}

In designing physical plausibility metrics, we draw on the Skating Ratio proposed in Gmd~\cite{gmd}, the Floating, Ground Penetration, and Foot Clipping metrics from ReinDiffuse~\cite{reindiffuse}, and the Speed and Smoothness metrics introduced by Fischer~\cite{motioneval}. The specific definitions of these metrics are as follows:

\begin{itemize}
    \item \textbf{Skating Ratio}: Foot skating refers to the situation where a person's foot is in contact with the ground but still has velocity. Given a human motion \(X\), the Foot Skating Ratio for the \(i\)-th frame can be expressed as:
    \begin{equation}
    FS^i(X) = \exp\left(-||\dot{X^i_{ft}} \cdot f^i_S \cdot f^i_V \cdot f^i_{\bar{V}} ||_2\right),
    \end{equation}
    where \(\dot{X^i_{ft}}\), \(f^i_S\), \(f^i_V\), and \(f^i_{\bar{V}}\) represent the foot velocity, the contact label, the instantaneous velocity threshold of the foot, and the average velocity threshold of the foot at the \(i\)-th frame, respectively.

    \item \textbf{Floating}: Floating refers to a scenario where the lowest joint of the generated motion is not in contact with the ground. Specifically, the Floating for the \(i\)-th frame can be represented as:
    \begin{equation}
    Float^i(X) = \exp\left(-||(X^i_h - h_{ground}) \cdot f^i_F ||_2\right),
    \end{equation}
    where \(X^i_h\) denotes the y-axis coordinate of the lowest point of the action at the \(i\)-th frame, \(h_{ground}\) is a threshold hyperparameter, set to 5cm in practice, and \(f^i_F\) is a flag that equals 1 when \(X^i_h > h_{ground}\) and 0 otherwise.

    \item \textbf{Ground Penetration}: Ground Penetration refers to the situation where the lowest joint of the generated motion penetrates below the ground. The Ground Penetration for the \(i\)-th frame can be expressed as:
    \begin{equation}
    GP^i(X) = \exp\left(-|| (h_{ground} - X^i_h) \cdot f^i_P ||_2\right),
    \end{equation}
    where \(f^i_P\) is a flag that equals 1 when \(X^i_h < h_{ground}\) and 0 otherwise.

    \item \textbf{Foot Clipping}: Foot Clipping refers to a situation where the two feet are too close together, causing mesh penetration. The Foot Clipping for the \(i\)-th frame can be expressed as:
    \begin{equation}
    FC^i(X) = \exp\left(-||(X^i_{lf} - X^i_{rf}) \cdot f^i_C ||_2\right),
    \end{equation}
    where \(X^i_{lf}\) and \(X^i_{rf}\) are the joint coordinates of the left and right feet, and \(f^i_C\) is a threshold function. \(f^i_C = 1\) when the distance between the two feet exceeds a threshold, which is set to 5cm in practice.
    
    \item \textbf{Speed}: Speed can be used to calculate the amplitude of human motion. The calculation for Speed at the \(i\)-th frame is given by:
    \begin{equation}
    Speed^i(X) = \frac{||X^i||_2}{\Delta_t},
    \end{equation}
    where \(\Delta_t\) is the sampling frame rate of the dataset, which is 0.05 for HumanML3D and 0.08 for KIT-ML.
    
    \item \textbf{Smoothness}: Smoothness serves as a measure of fluctuations in motion acceleration. Models of inferior quality often exhibit low smoothness in their generated motions. The calculation for Smoothness at the \(i\)-th frame is given by:
    \begin{equation}
    Smooth^i(X) = \frac{|\ddot{X^i} - \ddot{X^{i-1}}|}{\Delta_t}.
    \end{equation}
\end{itemize}

It is important to note that all the aforementioned metrics should ideally be as close as possible to the data set. Since the text-to-motion datasets include motions such as crawling, climbing stairs, and swimming, metrics like Skating Ratio and Floating are not necessarily better when smaller. For the KIT-ML dataset~\cite{kitml}, Ground Penetration and Clipping are also not zero, which may be related to the quality of the KIT-ML dataset.

\subsection{Quantitative Results}

\begin{figure}[t!]
    \centering
    \includegraphics[width=0.9\linewidth]{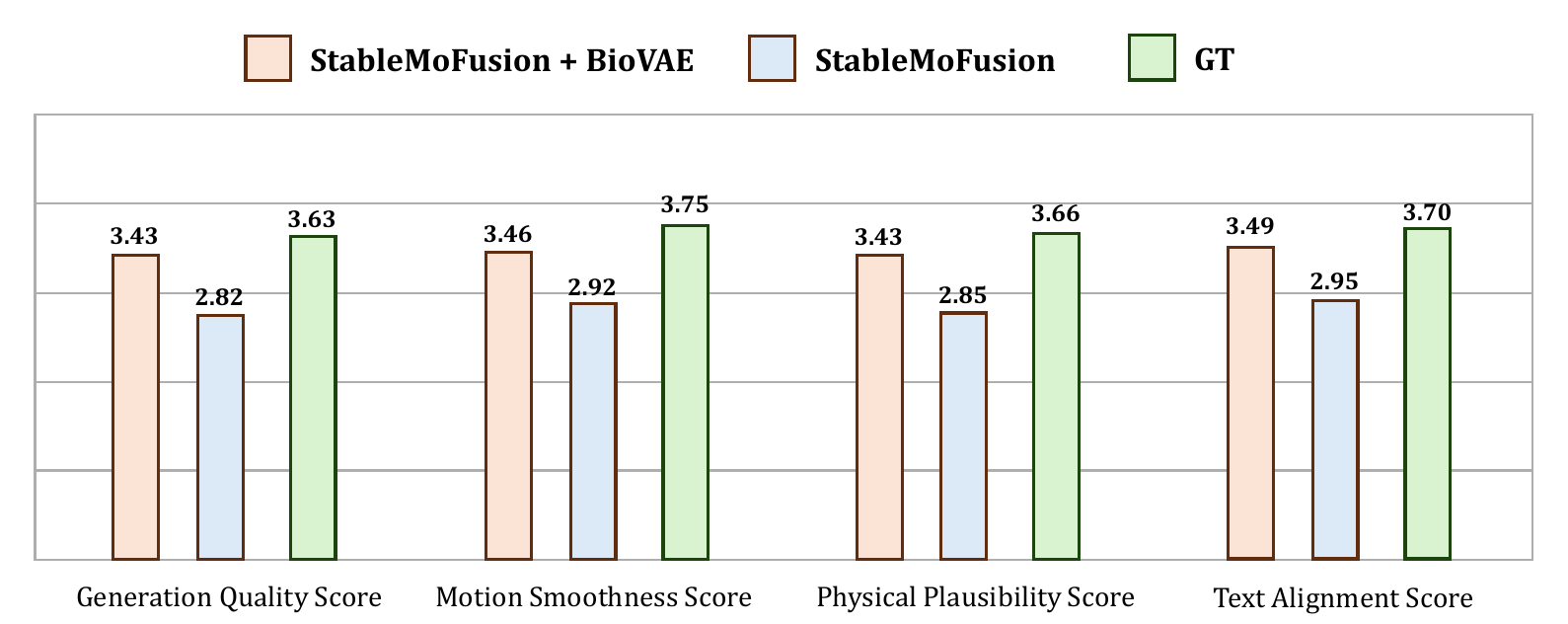}
    \caption{\small \textbf{User Study Results.}}
    \label{fig:user}
    %\vspace{-10pt}
\end{figure}

\paragraph{User Study.} We conducted a user study in which participants rated four aspects of the generated motions on a scale of 1 to 5: ``Generation Quality", ``Motion Smoothness", ``Physical Plausibility" and ``Text Alignment", using the HumanML3D test set. Thirty participants were invited to evaluate and compare the motion generation results of StableMoFusion + BioVAE, StableMoFusion, and the ground truth across 20 prompts. The results show that our model outperforms StableMoFusion in generation quality, motion smoothness, physical plausibility, and semantic alignment. In particular, our method is capable of generating results nearly indistinguishable from the ground truth in some cases, highlighting its effectiveness.

\begin{figure}[htbp]
    \centering
    \includegraphics[width=\linewidth]{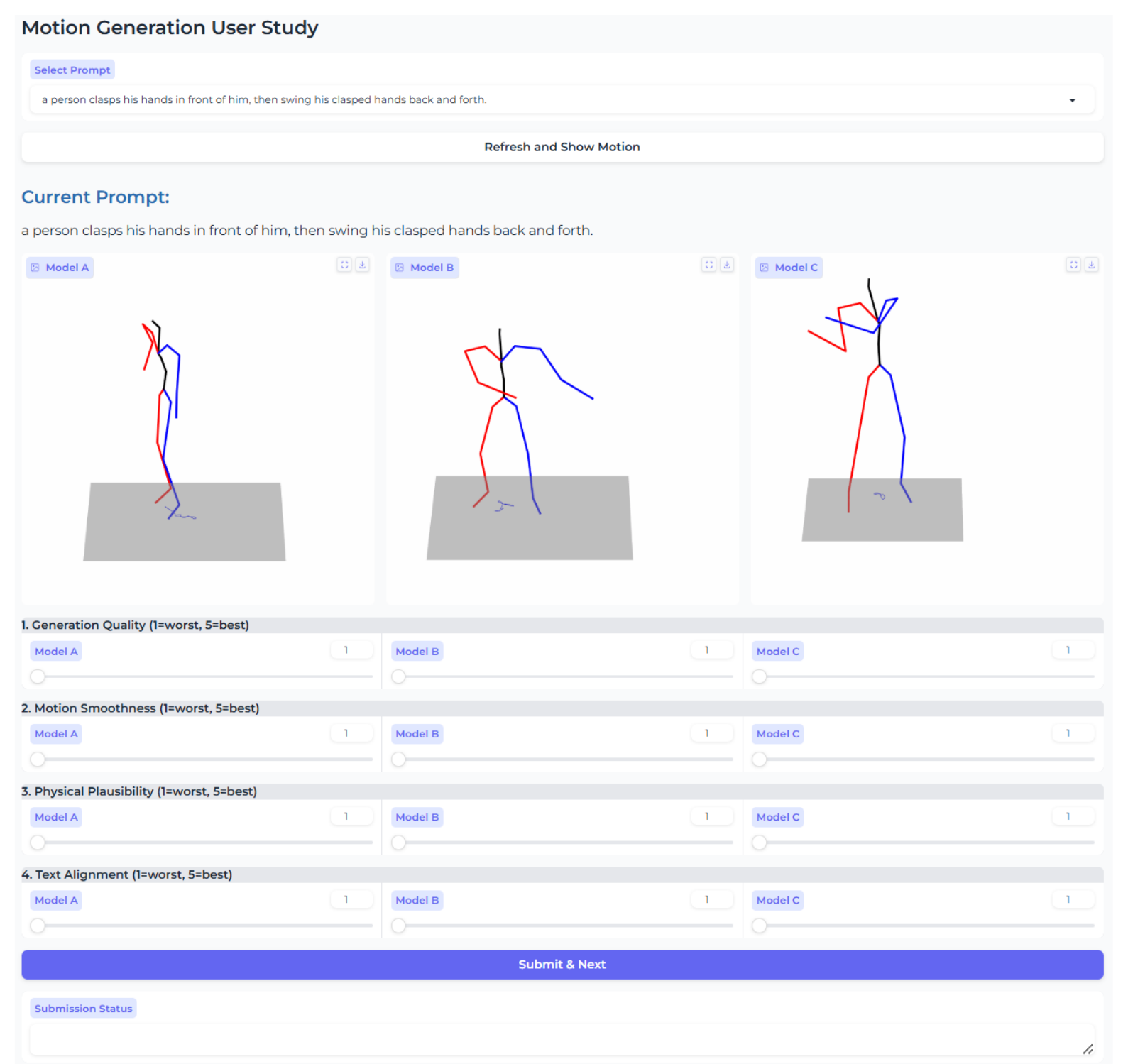}
    \caption{\small \textbf{User study interface for evaluating on HumanML3D.} To ensure unbiased feedback, the samples were presented in random order to avoid potential scoring inertia among participants.}
    \label{fig:userstudy}
    \vspace{-5pt}
\end{figure}

\paragraph{Main Results.} \Cref{tab:evaluator} and \Cref{tab:physics_evaluator} compare our framework with existing approaches in both physical plausibility metrics (e.g., smoothness, foot sliding) and traditional generation metrics (FID, Diversity). Our approach achieves state-of-the-art results in biomechanical authenticity while simultaneously advancing traditional motion quality, outperforming prior works across multiple dimensions. Experiments reveal critical limitations in existing paradigms: GPT-style generative models tend to over-smooth motions, failing to capture realistic dynamics, while diffusion-based methods exhibit temporal discontinuity, probably due to their implicit treatment of spatio-temporal relationships without explicit temporal modeling. In contrast, our method leverages the residual joint training of BioVAE and diffusion models, enabling the diffusion process to effectively capture acceleration information during training. Meanwhile, the incorporation of biomechanical priors reduces the complexity of the solution space, allowing our model to achieve superior performance on traditional metrics without compromising motion diversity.

\begin{figure*}[htbp]
    \centering
    \includegraphics[width=0.97\linewidth]{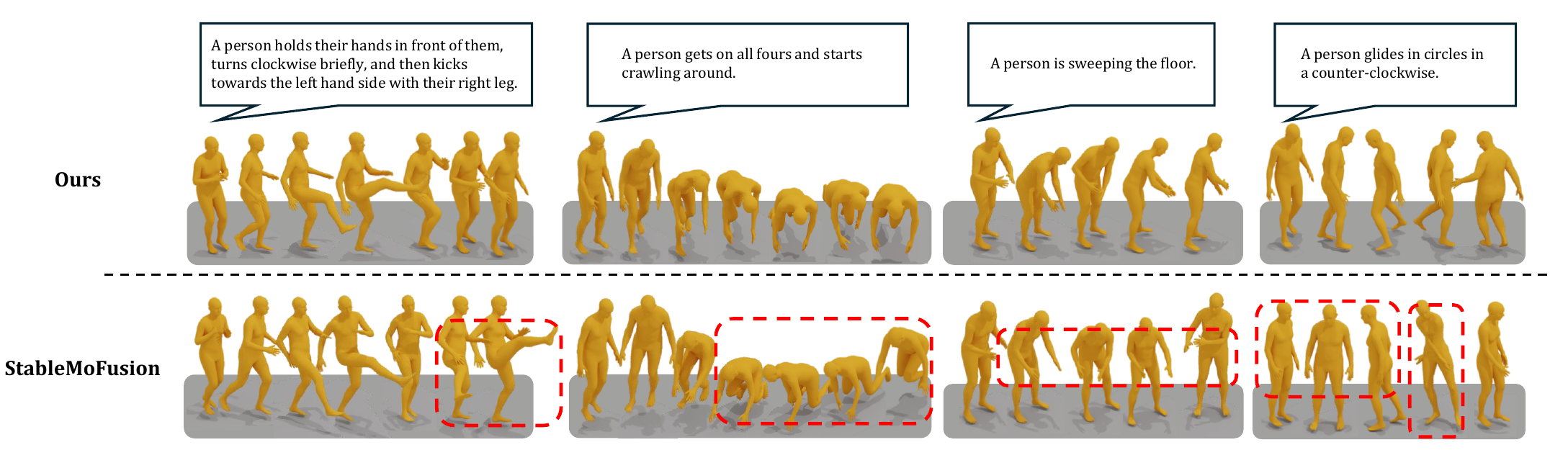}
    \caption{\small \textbf{Qualitative comparison of the state-of-the-art methods.} The results demonstrate that our method achieves superior language-motion alignment, enhanced temporal smoothness, and improved physical plausibility compared to current  approaches.}
    \label{fig:experiment}
    \vspace{-5pt}
\end{figure*}

\subsection{Ablation Study}

\begin{table}[t!]
\centering
\caption{\small \textbf{Ablation study on traditional metrics.} All experiments were repeated 20 times with different random seeds, and the mean results are reported in the tables.}
\label{tab:abl_gen}
\renewcommand\arraystretch{1.1}
\resizebox{\columnwidth}{!}{ % This command resizes the table to fit the column width
\begin{tabular}{ccccccccccc}
\hline
\multirow{2}{0.8cm}{Methods} & \multirow{2}{1.0cm}{BioVAE} & \multirow{2}{1.6cm}{MIA Pre-train} & \multirow{2}{1.6cm}{BioVAE Loss} & \multirow{2}{*}{FID$\downarrow$} & \multirow{2}{*}{MM Dist$\downarrow$} & \multirow{2}{*}{Div$\rightarrow$} & \multirow{2}{*}{MM$\uparrow$} & \multicolumn{3}{c}{R Precision$\uparrow$} \\
& & & & & & & & Top 1 & Top 2 & Top 3 \\
\hline
% GT     & -  & -  & -  & 0.002  & 2.974  & 9.503  & -  & 0.511  & 0.703  & 0.797  \\
% \hline
\multirow{4}{*}{Ours}  
    & $\times$  & $\times$  & $\times$  & $0.090$  & $2.824$  & $9.590$  & $1.849$  & $0.542$  & $0.739$  & $0.832$  \\
    ~& $\checkmark$ & $\times$ & $\times$ & $0.085$ & $\bm{2.760}$ & $9.764$ & $1.835$ & $\bm{0.551}$ & $\bm{0.747}$ & $\bm{0.839}$ \\
    ~& $\checkmark$  & $\checkmark$  & $\times$  & $\bm{0.066}$  & $2.783$  & $9.596$  & $\bm{1.948}$  & $0.549$  & $0.746$  & $0.836$  \\
    ~& $\checkmark$ & $\times$ & $\checkmark$ & $0.077$ & $2.803$ & $9.752$ & $1.873$ & $0.545$ & $0.740$ & $0.832$ \\
    ~& $\checkmark$  & $\checkmark$  & $\checkmark$  & $0.071$  & $2.784$  & \bm{$9.567$}  & $1.919$  & $0.547$  & $0.743$  & $0.835$  \\
\hline
\end{tabular}}
\end{table}

\begin{table}[t!]
\centering
\caption{\small \textbf{Ablation study on physical plausible metrics.}}
\label{tab:abl_phys}
\setlength{\tabcolsep}{0.9mm}
\renewcommand\arraystretch{1.2}
\resizebox{\columnwidth}{!}{ % This command resizes the table to fit the column width
\begin{tabular}{cccccccccccc}
\hline
Methods & BioVAE & MIA Pre-train & BioVAE Loss & VC$\rightarrow$ & AC$\rightarrow$ & Speed$\rightarrow$ & Smooth$\rightarrow$ & Float$\rightarrow$ & FS$\rightarrow$ & Pen$\rightarrow$ & Clip$\rightarrow$ \\
\hline
GT     & -    & -   & -  & $0.124$  & $0.682$  & $0.358$  & $2.60$  & $0.205$  & $0.057$  & $0.000$  & $0.000$  \\
\hline
\multirow{4}{*}{Ours}  
    & $\times$  & $\times$  & $\times$  & $0.292$  & $2.220$  & $0.448$  & $0.74$  & $0.234$  & $0.074$  & $0.043$  & $0.002$  \\
    & $\checkmark$ & $\times$ & $\times$ & $0.161$ & $1.205$ & $0.418$ & $1.23$ & $0.227$ & $0.056$ & $0.026$ & $0.002$ \\
    & $\checkmark$  & $\checkmark$  & $\times$  & $0.135$  & $0.954$  & $0.421$  & $1.39$  & $0.227$  & $0.060$  & $0.026$  & $0.002$  \\
    & $\checkmark$ & $\times$ & $\checkmark$ & $\bm{0.134}$ & $\bm{0.944}$ & $0.425$ & $1.33$ & $0.211$ & $0.056$ & $0.039$ & $0.002$ \\
    & $\checkmark$  & $\checkmark$  & $\checkmark$  & $0.135$  & $0.959$  & \bm{$0.385$}  & $\bm{1.45}$  & $\bm{0.204}$  & $\bm{0.056}$  & $\bm{0.013}$  & $\bm{0.002}$  \\
\hline
\end{tabular}}
\vspace{-5pt}
\end{table}

Our proposed BioVAE not only incorporates additional biomechanical priors but also achieves stable joint training by modifying the diffusion loss function. Ablation studies in identical training settings (\Cref{tab:abl_gen} and \Cref{tab:abl_phys}) demonstrate the critical contributions of the BioVAE architecture, the biomechanical priors, and the newly introduced loss term. Specifically, BioVAE substantially improves motion generation performance on traditional metrics, while the biomechanical priors and BioVAE loss each independently enhance the physical plausibility of the generated motions. When combined, these components exhibit even greater improvements, highlighting the effectiveness and advancement of our framework.

\begin{figure*}[t!]
    \centering
    \includegraphics[width=0.85\linewidth]{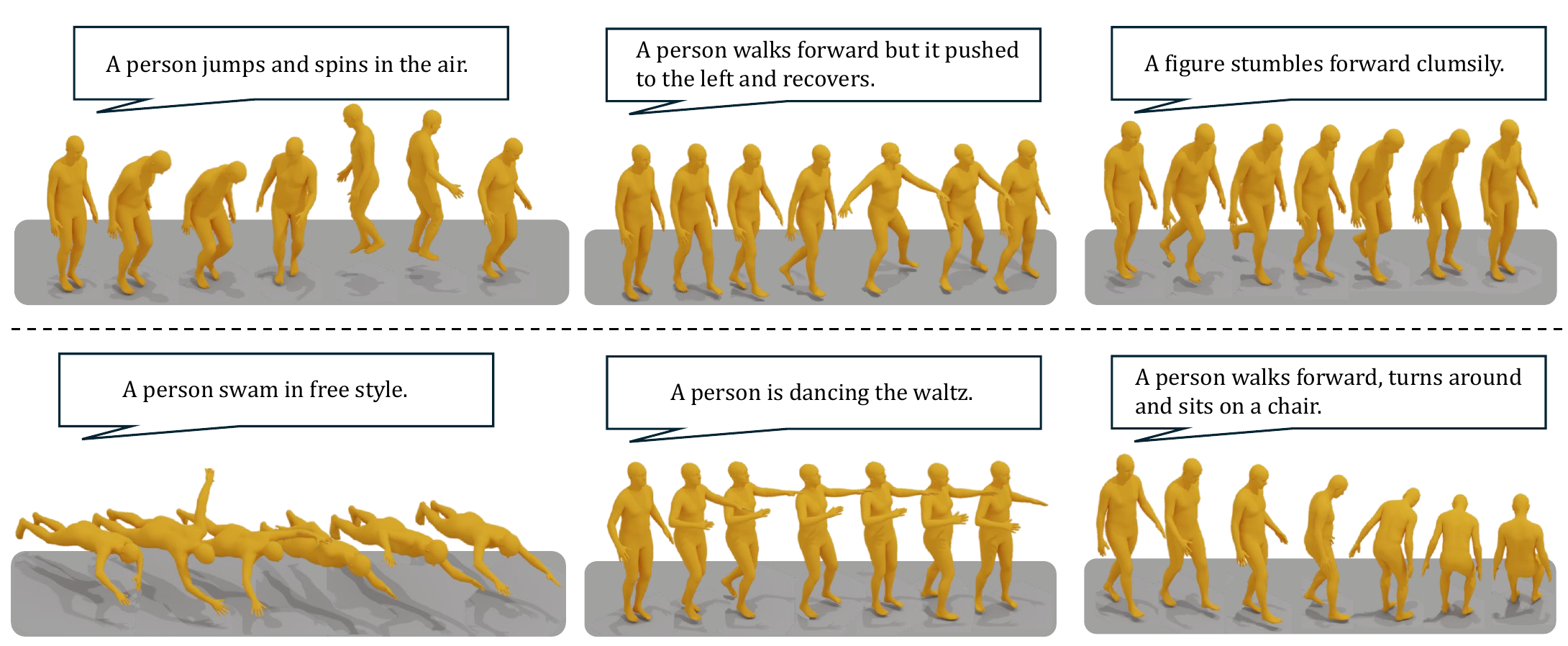}
    \caption{\small Additional samples from our text-to-motion synthesis approach, generated with text prompts from the HumanML3D test set. The results demonstrate that our approach effectively understands diverse textual semantics and generates physically plausible motions.}
    \label{fig:ourresult}
    \vspace{-10pt}
\end{figure*}

\subsection{Motion Estimation and Editing}

\begin{table}[htbp]
\vspace{-3pt}
\centering
\caption{\small \textbf{Comparative results for pose estimation.} We used WHAM~\cite{wham} and various diffusion models as the base framework. Building on the official WHAM implementation, we evaluated the impact of incorporating the residual-based BioVAE on pose estimation performance using the 3DPW~\cite{3DPW} and RICH~\cite{RICH} datasets. Bold numbers denote the most accurate method in each column. Accel is in $m/s^2$, all other errors are in $mm$.} 
\label{tab:pose_comparison}
\setlength{\tabcolsep}{1.4mm}
\renewcommand\arraystretch{1.1}
\resizebox{\columnwidth}{!}{
\begin{tabular}{lcccc|cccc}
\hline
Method & \multicolumn{4}{c|}{3DPW} & \multicolumn{4}{c}{RICH} \\
\cline{2-5} \cline{6-9}
 & PA-MPJPE$\downarrow$ & MPJPE$\downarrow$ & PVE$\downarrow$ & Accel$\downarrow$
 & PA-MPJPE$\downarrow$ & MPJPE$\downarrow$ & PVE$\downarrow$ & Accel$\downarrow$ \\
\hline
WHAM & 36.34 & 61.46 & 70.64 & 6.52 & 50.67 & 83.00 & 95.32 & 5.35 \\
MDM Inversion & 36.48 & 62.06 & 71.36 & 6.41 & 51.76 & 83.62 & 96.53 & 5.28 \\
StableMoFusion Inversion  & $\bm{36.06}$ & 61.85 & 71.26 & 6.29 & 51.64 & 83.12 & 96.53 & 5.11 \\
StableMoFusion + BioVAE Inversion & 36.13 & $\bm{61.11}$ & $\bm{70.52}$ & $\bm{5.55}$ & $\bm{50.60}$ & $\bm{81.80}$ & $\bm{94.84}$ & $\bm{4.06}$ \\
\hline
\end{tabular}}
\vspace{-5pt}
\end{table}

As shown in \Cref{tab:pose_comparison}, we evaluate the capability of our framework for motion refinement and pose estimation using WHAM~\cite{wham}, a state-of-the-art method in the pose estimation domain. WHAM incorporates sequential modeling and temporal information, achieving leading results in acceleration-related metrics. Our results demonstrate that integrating BioVAE with diffusion models and pose estimation frameworks can significantly improve the smoothness of human motion while also enhancing pose estimation accuracy, achieving state-of-the-art performance across all evaluated metrics.  

These experiments indicate that our physics-aware network not only reduces the complexity of the solution space and enhances the physical plausibility but also substantially improves the smoothness of the motion and the overall performance even without relying on acceleration loss supervision. This highlights the versatility of our approach to simultaneously refining human motion and advancing pose estimation results.

\begin{table}[t!]
\centering
\caption{\small \textbf{Motion Editing via Acceleration Modification.} We manually edited the acceleration values in BioVAE to evaluate the effects of increasing or decreasing acceleration on the generated motions. All experiments were conducted on the HumanML3D dataset.}
\label{tab:acc_weight}
\renewcommand\arraystretch{1.1}
\resizebox{\columnwidth}{!}{
\begin{tabular}{cccccccccccc}
\hline
\textbf{Acc} & \textbf{FID}$\downarrow$ & \textbf{MM Dist}$\downarrow$ & \textbf{RP3}$\uparrow$ & \textbf{VC}$\rightarrow$ & \textbf{AC}$\rightarrow$ & \textbf{Speed}$\rightarrow$ & \textbf{Smooth}↓ & \textbf{Float}$\rightarrow$ & \textbf{FS}$\rightarrow$ & \textbf{Pen}$\rightarrow$ & \textbf{Clip}$\rightarrow$ \\
\hline
GT & $0.002$ & $2.974$ & $0.797$ & $0.124$ & $0.682$ & $0.358$ & $2.60$ & $0.205$ & $0.057$ & $0.000$ & $0.000$ \\
\hline
$0.5$ & $0.479$ & $2.860$ & $0.818$ & $0.148$ & $1.085$ & $0.294$ & $1.83$ & $0.096$ & $0.070$ & $0.012$ & $0.001$ \\
$0.8$ & $0.114$ & $2.782$ & $0.834$ & $0.138$ & $0.989$ & $0.356$ & $1.56$ & $0.165$ & $0.061$ & $0.013$ & $0.002$ \\
$1.0$ & $0.071$ & $2.784$ & $0.835$ & $0.135$ & $0.959$ & $0.385$ & $1.45$ & $0.204$ & $0.056$ & $0.013$ & $0.002$ \\
$1.5$ & $0.085$ & $2.824$ & $0.830$ & $0.134$ & $0.937$ & $0.428$ & $1.32$ & $0.276$ & $0.050$ & $0.013$ & $0.002$ \\
$2.0$ & $0.124$ & $2.869$ & $0.824$ & $0.136$ & $0.942$ & $0.447$ & $1.26$ & $0.321$ & $0.045$ & $0.013$ & $0.002$ \\
\hline
\end{tabular}}
\vspace{-5pt}
\end{table}

By modifying the acceleration values in BioVAE for ablation studies, we observed the following findings: (1) Traditional generation metrics generally do not improve with changes in acceleration magnitude, which explains why generative models require parameter consistency between training and testing; (2) Certain physical metrics, such as ground penetration, foot clipping, and velocity/acceleration consistency, also show minimal sensitivity to acceleration changes; (3) Some physical metrics exhibit a strong correlation with acceleration magnitude: smoothness and skating ratio decrease as acceleration increases, while ground floating increases with higher acceleration values.  

We attribute these phenomena to the explicit representation of acceleration in our model. When we modify the absolute magnitude of the input acceleration, the model shows little impact on traditional generation metrics. However, because our method explicitly models acceleration, altering its absolute magnitude affects the generated motion amplitude and, in turn, influences acceleration-related physical plausibility metrics. This highlights the advantage of our approach: the ability to regulate acceleration to control the physical properties of generated motions. In addition, these insights provide guidance for optimizing specific physical metrics by tuning acceleration-related parameters.  

\subsection{Implementation Details}

All experiments were conducted on a single NVIDIA A40 GPU. For our implementation of motion generation, we adopt StableMoFusion as the baseline. We employ DDPM~\cite{ddpm} with 1,000 denoising steps (\(T = 1{,}000\)) and linearly varying variances \(\beta_t\) from 0.0001 to 0.02 in the forward process. The model is trained using AdamW~\cite{adamw} with an initial learning rate of 0.0002 and a weight decay of 0.01 for 50{,}000 iterations during pre-training, followed by 10{,}000 iterations of fine-tuning with a batch size of 64. Additionally, we apply an Exponential Moving Average (EMA) to smooth parameter updates and reduce instability and oscillations. During inference, we utilize the SDE variant of the second-order DPM-Solver++~\cite{dpmsolver} with Karras sigmas~\cite{karras} for 10-step sampling. We set the CFG scale for text conditioning to 2.5.

For our implementation of pose refinement, we adopted WHAM~\cite{wham} as the baseline. However, the official WHAM code repository provides models trained only on five datasets: AMASS~\cite{AMASS}, InstaVariety~\cite{InstaVariety}, MPI-INF-3DHP~\cite{3dhp}, Human3.6M~\cite{h36m}, and 3DPW~\cite{3DPW}, and does not include a version trained with the BEDLAM~\cite{BEDLAM} dataset. Consequently, the baseline results reported in our work differ slightly from those presented in the original WHAM paper.

\begin{figure}[t!]
    \centering
    \includegraphics[width=\linewidth]{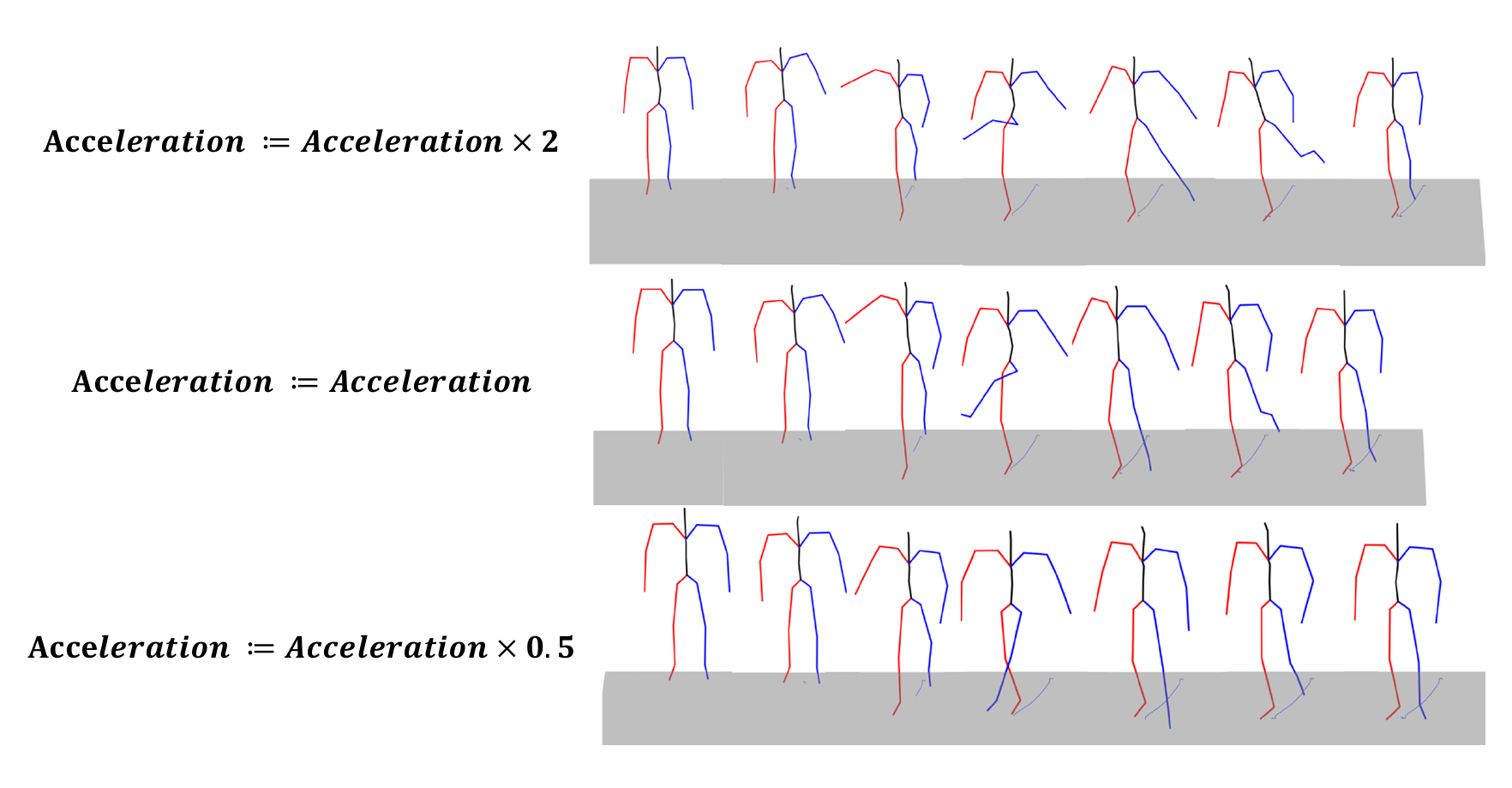}
    \caption{\small \textbf{Editing generated human motions to different movement amplitudes.} By manually modifying the acceleration values in BioVAE, controllable motion generation can be achieved. Adjusting the acceleration for a single joint or all joints allows the generated motions to exhibit varying amplitudes.}
    \label{fig:speed}
    \vspace{-15pt}
\end{figure}

\section{Conclusion}
\label{sec:conclusion}

In this paper, we propose BioVAE, a biomechanics-aware framework designed for a wide range of tasks, including motion generation, pose estimation, motion refinement, and motion editing. BioVAE is inspired by extended Euler-Lagrange equations and achieves deep integration with the diffusion process through a carefully designed loss function. Furthermore, it incorporates joint learning from electromyographic (EMG) signals and acceleration data. Experimental results demonstrate that our approach outperforms previous models in both generation quality and physical plausibility. However, data limitations have so far prevented us from evaluating our approach on high-frame-rate datasets with richer annotations. Addressing this limitation in future work could enable BioVAE to deliver even stronger results for practical industrial applications.  

%  {\appendix[Proof of the Zonklar Equations]
% Use $\backslash${\tt{appendix}} if you have a single appendix:
% Do not use $\backslash${\tt{section}} anymore after $\backslash${\tt{appendix}}, only $\backslash${\tt{section*}}.
% If you have multiple appendixes use $\backslash${\tt{appendices}} then use $\backslash${\tt{section}} to start each appendix.
% You must declare a $\backslash${\tt{section}} before using any $\backslash${\tt{subsection}} or using $\backslash${\tt{label}} ($\backslash${\tt{appendices}} by itself
%  starts a section numbered zero.)}

% {\appendices
% \section*{Proof of the First Zonklar Equation}
% \section*{Proof of the Second Zonklar Equation}
% Appendix two text goes here.}

\bibliographystyle{IEEEtran}
\bibliography{ieeetmm}

\end{document}